\documentclass[fleqn,10pt]{wlscirep}
\usepackage[utf8]{inputenc}
\usepackage[T1]{fontenc}
\pdfoutput=1
\usepackage{graphicx}
\usepackage{lineno}
\title{Estimating and Abstracting the 3D Structure of Bones Using Neural Networks on X-Ray (2D) Images}

\author[1]{Jana Čavojská}
\author[1]{Julian Petrasch}
\author[1]{Nicolas J. Lehmann}
\author[1]{Agnès Voisard}
\author[2]{Peter Böttcher}
\affil[1]{Freie Universität Berlin, Institute of Computer Science, Berlin, 14195, Germany}
\affil[2]{Freie Universität Berlin, Clinic for Small Animals, Berlin, 14163, Germany}

\begin{abstract}
In this paper, we present a deep-learning based method for estimating the 3D structure of a bone from a pair of 2D X-ray images. Our triplet loss-trained neural network selects the most closely matching 3D bone shape from a predefined set of shapes. Our predictions have an average root mean square (RMS) distance of $1.08 \, mm$ between the predicted and true shapes, making it more accurate than the average error achieved by eight other examined 3D bone reconstruction approaches. The prediction process that we use is fully automated and unlike many competing approaches, it does not rely on any previous knowledge about bone geometry. Additionally, our neural network can determine the identity of a bone based only on its X-ray image. It computes a low-dimensional representation (``embedding'') of each 2D X-ray image and henceforth compares different X-ray images based only on their embeddings. An embedding holds enough information to uniquely identify the bone CT belonging to the input X-ray image with a 100\% accuracy and can therefore serve as a kind of fingerprint for that bone. Possible applications include faster, image content-based bone database searches for forensic purposes.

\end{abstract}
\begin{document}

\flushbottom
\maketitle

\thispagestyle{empty}


\section*{Introduction}

Traditional methods for reconstructing 3D bone models from Computed Tomography (CT) scans require high-radiation dose, cost and time. For this reason, generating 3D models in a different way, for example directly from 2D images, can be a useful alternative \cite{3Dlaplace_peter}. Currently, multiple approaches for estimating the 3D structure of bones from their 2D X-ray images exist 
 \cite{3Dlaplace_peter}\hspace{0.05cm} 
 \cite{SSM_baka}\hspace{0.05cm} 
 \cite{FFD_filippi}\hspace{0.05cm} 
 \cite{SSM_fleute}\hspace{0.05cm} 
 \cite{nonSSM_gamage}\hspace{0.05cm} 
 \cite{nonSSM_laport}\hspace{0.05cm} 
 \cite{nonSSM_tang}\hspace{0.05cm} 
 \cite{SSM_zhu} 
 which help reduce the cost and the radiation-related health risks for the patient. There are even a few fully automated approaches \cite{galibarov}\hspace{0.05cm} \cite{prakoonwit}. However, the first fully automated approach that we found requires previous knowledge about the bone geometry in order to identify bone boundaries in the input image \cite{galibarov}, and the second approach requires five X-ray images taken from different angles \cite{prakoonwit}. A fully automated approach which can easily adapt to new data without previous knowledge about bone geometry or other attributes is therefore desirable.

We present a fully automated neural network-based method for estimating the 3D structure of a bone, given its 2D X-ray scan. Our method is completely domain-agnostic, which means that all that is needed for teaching the network to work with completely different bones, such as human tibiae instead of cat femurs, is a set of 3D CT images of such new bones. Our method is based on assigning the most closely matching 3D shape of a bone to the 2D input image of that bone by selecting the shape match from a pre-existing set of 3D shapes. We treat the search for an optimal 3D shape as a classification problem. Each class corresponds to one specific 3D shape and is presented to the classifier during training as a set of 2D images generated from the 3D shape. To build a classifier that solves this problem, we trained a convolutional neural network (CNN) using the triplet loss 
\cite{triplet_network} \hspace{0.05cm}
\cite{facenet} \hspace{0.05cm}
\cite{person_reident} \hspace{0.05cm}
\cite{triplet_Cheng_2016_CVPR} \hspace{0.05cm}
\cite{triplet_G_2016_CVPR} \hspace{0.05cm}
\cite{triplet_Zhuang_2016_CVPR} \hspace{0.05cm}
\cite{triplet_He_2018_CVPR} \hspace{0.05cm}
\cite{triplet_bui2017compact} \hspace{0.05cm}
\cite{triplet_liu2016multi} \hspace{0.05cm}
method and taught it to differentiate between femurs of different cats based on artificial 2D X-ray images generated from the 3D femur CT scans. Once fully trained, the network was able to generate for each input image a low-dimensional representation of its content, a so-called $d$-dimensional image embedding. We then trained a k-nearest neighbor (kNN) classifier on these embeddings. To predict the shape of a new bone, the kNN classifier assigned the 2D input image embedding of that bone its closest embedding of an image from the training set. Since the 3D shapes of the bone images from the training set are known, this assignment results in identifying the 3D shape that most closely matches the bone in the input image, as determined by the properties (features) extracted by the neural network.

How well these properties, which were considered relevant by the network, correlate with the bones' actual 3D shapes was evaluated by computing the root mean square (RMS) and Hausdorff distances \cite{metro}\hspace{0.05cm} \cite{hausdorff_dist_paper} between sample bones and the ground truth (the actual 3D shape of the sample bone). Neither the network nor the kNN classifier were trained on the sample bones. To put these evaluation results into context, the measured distances were also compared with the distances obtained using statistical shape models (SSM) for 3D shape reconstruction of the same bones. The evaluation results were also compared with distances achieved on similar tasks in related literature.

In addition, we discovered that training a neural network using triplet loss results in the network's ability to generate embeddings (vectors) that can be used to uniquely identify the specific bone the input image depicts. These embeddings are easily separable by their Euclidean distance, even in case of bones that were never presented to the network in any way during training. Embeddings of artificial X-ray images generated from the same CT scan build tight groups in the Euclidean space, while embeddings from different CTs and hence different bones are distinctly further apart, which has interesting implications for data compression purposes and for searching bone databases.

In this paper, we first briefly describe our dataset, then we present the results achieved using a pre-trained, generic neural network classifier. After that, we present the approach of separating our feature extractor from the classifier and the triplet loss method we used. We show how the extracted features can be used not only for classification, but also for pairwise comparisons of the 3D CT volumes the X-ray images originated from, based on the uniqueness of the X-ray image fingerprints. We finally evaluate the triplet loss approach to bone shape estimation qualitatively by showing alignments between our predictions and the true shapes, quantitatively by comparing the achieved RMS distance between our predictions and the true shapes with the RMS distance achieved using statistical shape models on our data, and also by comparing with the error achieved by different methods found in other literature.


\section*{Results}

\subsubsection*{Dataset}

\begin{figure}[ht]
\centering
\includegraphics[width=\linewidth]{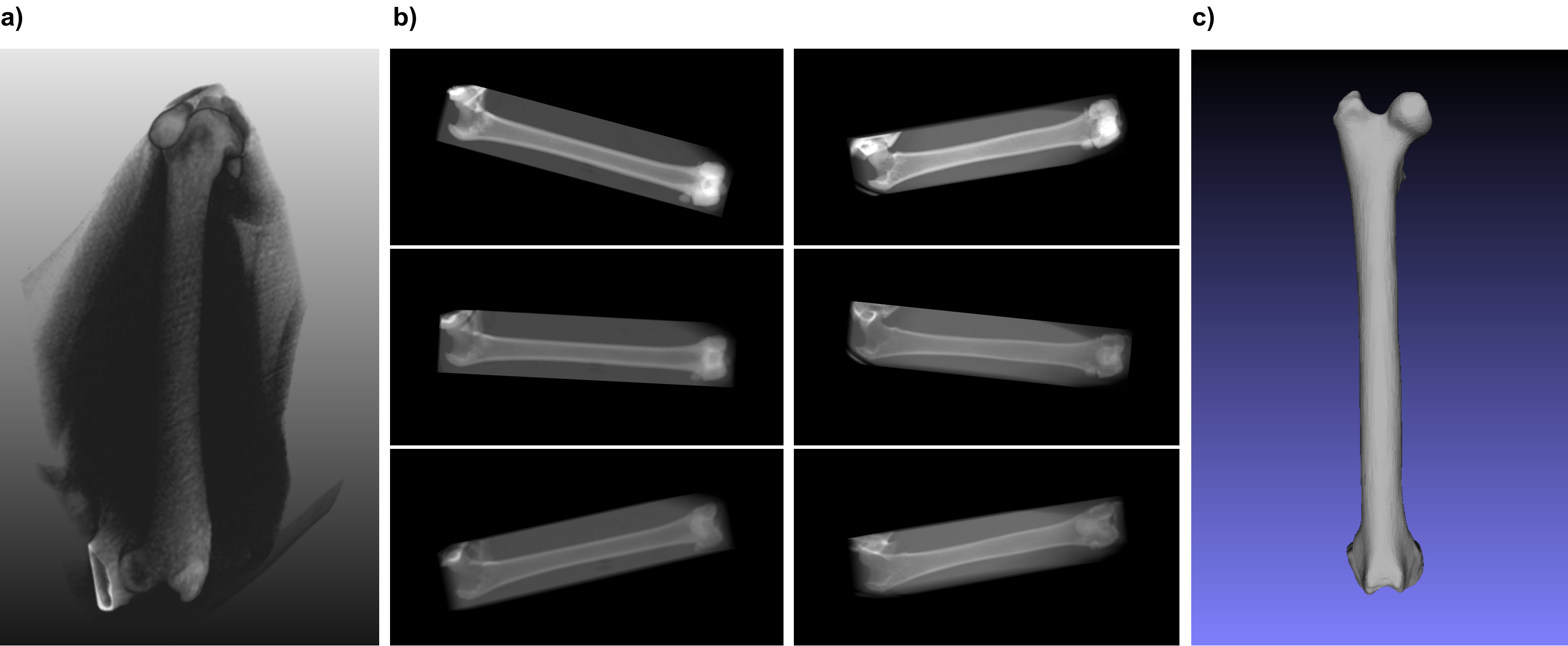}
\caption{Dataset. \ref{fig:dataset}a) 3D CT DICOM file. \ref{fig:dataset}b) examples of X-ray images artificially generated from 3D CT DICOM data. Images in the left column of \ref{fig:dataset}b) were generated from the same bone. Images in the right column of \ref{fig:dataset}b) were generated from a second bone. \ref{fig:dataset}c) bone mesh (surface model) extracted from the 3D CT DICOM file}
\label{fig:dataset}
\end{figure}

Our dataset consisted of 29 3D CT scans of 29 femurs of different cats in the DICOM format. Later, three new CT scans were added, each paired with two natural X-ray scans of the same bone as in the CT scan; the first natural X-ray scan showed the bone from the anteroposterior, the second from the mediolateral view. Using the software \textit{MeVisLab}, we generated 900 artificial 2D X-ray images for each of the 29 CT scans by varying the viewing angle and radiation amount, thus simulating the variability of conditions during natural X-ray scans \cite{xray_positioning_variability}\hspace{0.05cm} \cite{radiation_body_weight}. These 26100 images served as the dataset to train and evaluate the neural network on. For evaluation purposes, the 3D shape in the STL format was extracted from each 3D CT scan so that the difference between a predicted shape and the true shape could be measured. The different data types are shown in Figure \ref{fig:dataset}.

\subsubsection*{Classification vs. feature extraction using a neural network}

We first tested neural networks' ability to extract meaningful features from 2D X-ray images of bones by performing transfer learning \cite{midlevel_transfer} on the dataset of 26100 images of cat femurs artificially generated from 3D CT scans. Different network architectures pre-trained on the ImageNet \cite{imagenet} benchmark dataset were used. The last layer of these networks, the dataset-specific softmax layer \cite{transfer_learning_softmax}, was removed and replaced by a softmax layer specific to our bone image dataset. The network, modified this way, was then further trained on our bone image dataset. It made no difference for the achieved accuracy whether only the new softmax layer was trained or the earlier layers were fine-tuned, as well. Implementations using different framework and neural network combinations were tested:
\begin{itemize}
    \item Keras with the ResNet-50 \cite{resnet} network and a validation accuracy of 86.13\% 
    \item Keras with the VGG-16 \cite{vgg} network and a validation accuracy of 63.13\% 
    \item TensorFlow with the Inception-ResNet-V2 network and a validation accuracy of 85.80\% 
    \item TensorFlow Hub with the ResNet-50 or Inception-V3 networks and a validation accuracy of 100.00\%
\end{itemize}
Neither different optimizers (Stochastic Gradient Descent, Adam) nor different learning rates (0.01, 0.0001) made a difference in accuracy greater than 1\%.

The transfer learning experiments showed that neural networks have the ability to classify bones based on their artificial X-rays with a high accuracy, 100\% in case of the TensorFlow Hub implementation. Such transfer learning classification would have been sufficient to estimate the 3D structure of bones: A class would simply be inferred by the network for the bone X-ray input image. Such a class would represent the most closely matching bone shape. However, additional functionality was achieved by training the neural network as a pure feature extractor instead of performing transfer learning. This feature extractor approach was inspired by publications about face classification \cite{facenet} and person re-identification \cite{person_reident}.

\subsubsection*{Attention map visualization of the classes after transfer learning}

In order to verify whether the networks used for transfer learning were capable of extracting the right features, we applied the grad-CAM (Gradient-weighted Class Activation Mapping) \cite{gradCAM} visualization to images from the 29 classes. The grad-CAM visualization approach identifies the regions of the input image which are most relevant for predicting a certain class. It does so by analyzing the gradients flowing into the last convolutional layer of a network. Specifically, in order to find the regions of the image most relevant to a specific class, Selvaraju et al. compute the gradient of the score for that class with respect to the feature maps of the last convolutional layer. The last convolutional layer is chosen for this purpose because it has been shown that deeper network layers capture higher-level visual constructs (such as whole objects, as opposed to simple edges or textures). And unlike fully connected layers, convolutional layers of a network retain the spatial information from the input image which makes it possible for grad-CAM to localize the most relevant complex features in the input image \cite{gradCAM}.

\begin{figure}[ht]
\centering
\includegraphics[width=\linewidth]{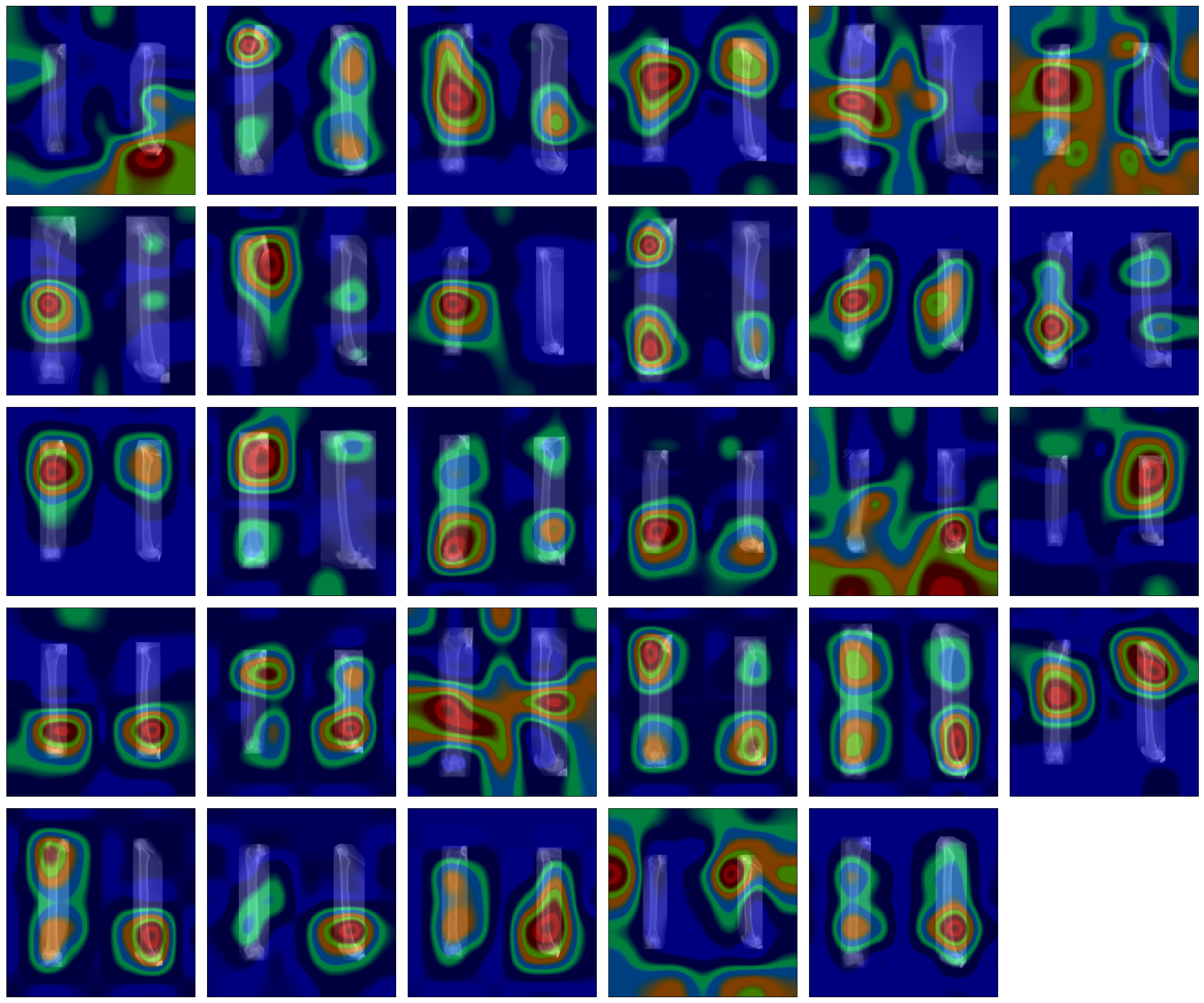}
\caption{Attention maps overlaid over the bone images of the 2D image dataset. Each of the 29 images shows one of the cat femurs from the anteroposterior and mediolateral view next to each other, overlaid by a heatmap which highlights in bright red the regions of the image that were most relevant for its correct classification using the VGG-16 network.}
\label{fig:attention-maps}
\end{figure}

Figure \ref{fig:attention-maps} shows the results of the grad-CAM visualization applied to the VGG-16 network after transfer learning. Each blue square represents one of the 29 bones, and the bright red overlays highlight the regions of an image which were most relevant for its correct classification. On almost all the images (except for the penultimate one), the bright red region coincides with a part of the bone. In most images, both bones are being taken into consideration. This visualization has not shown any obvious defects in the dataset, such as the network identifying artifacts (such as the crop boxes around the bones) or any other unexpected anomalies which could be easily remedied. Even in the cases where a big portion of the relevant regions lies outside of both bones, the most relevant region still overlaps with the bones. This is a coarse indication that the features extracted by the network do not have any obvious defects and augmenting the dataset further is not necessary.

\subsubsection*{Feature extraction using triplet loss}

In order to be able to use the features extracted from the 2D X-ray images for a variety of purposes, not only classification, we separated the feature extraction step from the classification step in a way also used in face recognition and person re-identification publications \cite{facenet}\hspace{0.05cm} \cite{person_reident}.

We modified the network architecture ResNet-50 \cite{resnet} and later VGG-16 \cite{vgg} and incorporated these architectures, in consecutive experiments, into a Triplet network architecture which was then trained with the triplet loss method. When performing transfer learning, the neural network is trained by iteratively being shown examples of each class, computing the error (difference between predicted class and true class) and modifying the network parameters to decrease this error. The Triplet network was instead trained by being iteratively shown groups of three images, so-called triplets: the first image of a triplet, so-called \textit{anchor ($x_i^a$)}, was selected randomly. The second image, \textit{positive ($x_i^p$)}, was selected from the same class as \textit{anchor}. The third image, \textit{negative ($x_i^n$)}, was selected from a different class than \textit{anchor}. The error was computed as follows: For each of the triplet images, its low-dimensional representation ($d$-dimensional embedding) was computed. The embeddings were computed by the \textit{base network} part of the 
Triplet network, consisting of either ResNet-50 or VGG-16, followed by dimension reduction layers. Each embedding was normalized to have unit-length, i.e. the embeddings were forced to live on the surface of a $d$-dimensional hypersphere. The Euclidean distances between the embeddings of \textit{anchor} and \textit{positive}, as well as between the embeddings of \textit{anchor} and \textit{negative}, were computed. Like in the work of Schroff et al. \cite{facenet}, the loss $L$ to be minimized, i.e. the ``triplet loss'', was then computed as 

\begin{equation}
    L = \sum_i^N{max\left( 0,\; \|f(x_i^a) - f(x_i^p)\|_2^2 \; - \; \|f(x_i^a) - f(x_i^n)\|_2^2 \; + \; margin\right)} 
    \label{eq:triplet_loss}
\end{equation}

where
\begin{itemize}
    \item $f(x_i^a)$ is the network's representation ($d$-dimensional embedding) of the \textit{anchor} image. The embedding dimension, $d$, is selected prior to training.
    \item $f(x_i^p)$ is the embedding of the \textit{positive} image
    \item $f(x_i^n)$ is the embedding of the \textit{negative} image
    \item $\|f(x_i^a) - f(x_i^p)\|_2^2$ is the squared Euclidean distance between the embeddings of the \textit{anchor} and the \textit{positive} image
    \item $\|f(x_i^a) - f(x_i^n)\|_2^2$ is the squared Euclidean distance between the embeddings of the \textit{anchor} and the \textit{negative} image
    \item \textit{margin} is the minimum Euclidean distance between embeddings of different classes enforced by the triplet loss function during training
\end{itemize} 
This loss function was designed to minimize the Euclidean distance 
between the \textit{anchor} embedding and the \textit{positive} embedding while maximizing the distance between the \textit{anchor} embedding and the \textit{negative} embedding. The loss function also enforces a minimum distance of \textit{margin} between the embeddings of different classes. It did not make any difference in accuracy whether or nor the Euclidean distance was squared.\\
\\
The triplet accuracy function was then defined as the percentage of triplets for which the following condition was met:

\begin{equation}
    \|f(x_i^a) - f(x_i^p)\|_2^2 \; + margin \; < \; \|f(x_i^a) - f(x_i^n)\|_2^2 \;
    \label{eq:triplet_accuracy}
\end{equation}

The Triplet network accuracy was computed as the percentage of triplets for which the Euclidean distance between same-class image embeddings was smaller, by \textit{margin}, than the Euclidean distance between different-class embeddings.

The optimal value for the hyperparameter \textit{margin} was experimentally determined to be $0.1$ for the bone dataset. The value $1.0$ lead to an accuracy drop of $9.6\%$, making it the the parameter which influenced the accuracy most.

In related work \cite{facenet}\hspace{0.05cm} \cite{audioset_paper}, $d$ is set to $128$. We tested the embedding dimensions 128, 64, 32, and 16. An accuracy drop ($~1\%$) was only noted with 16-dimensional embeddings. After these tests, we conducted all experiments with $d=128$.

The validation accuracy of the Triplet network achieved $99.9\%$ when the following options were used: a \textit{margin} value of $0.1$, using 900 images per class, X-ray images showing bone from anteroposterior next to a mediolateral view, 128-dimensional embeddings, using the VGG-16 network rather than ResNet-50, and using X-ray images with bones scaled relatively to their real-world proportions. The choice of the hyperparameter\textit{margin} had by far the biggest impact. The second most important option after \textit{margin} was a sufficient number of dataset images; when the number was decreased to 100 images per class, a validation accuracy drop of $3.5\%$ was observed. Using images with the combined anteroposterior and mediolateral view increased accuracy by $2.0\%$ as opposed to only using anteroposterior images. Using variable learning rate as opposed to a constant one for all layers improved the accuracy by $1.0\%$. Replacing ResNet by VGG as part of the base network within the Triplet network improved the accuracy by $0.8\%$. Scaling bones according to their real-world proportions improved the accuracy by $0.7\%$.

The following options had no effect on the Triplet accuracy: the specific layers used to achieve dimension reduction at the end of the base network, using the tertiary distance to compute loss (the distance between embeddings of \textit{positive} and \textit{negative} images), and reducing the number of classes from 29 to 24. 

The Triplet network trained this way has the ability to generate embeddings of its input images which have the property that the dissimilarity between images directly translates into the Euclidean distance between their embeddings. The more dissimilar the images, the higher the Euclidean distance between their embeddings.

The triplet accuracy of $99.9\%$ means that the network has learned very well to group together images in the Euclidean space that belong to the same classes, while keeping them apart from images belonging to different classes. This property made it possible to use the embeddings for classification purposes.

\begin{figure}[ht]
\centering
\includegraphics[width=\linewidth]{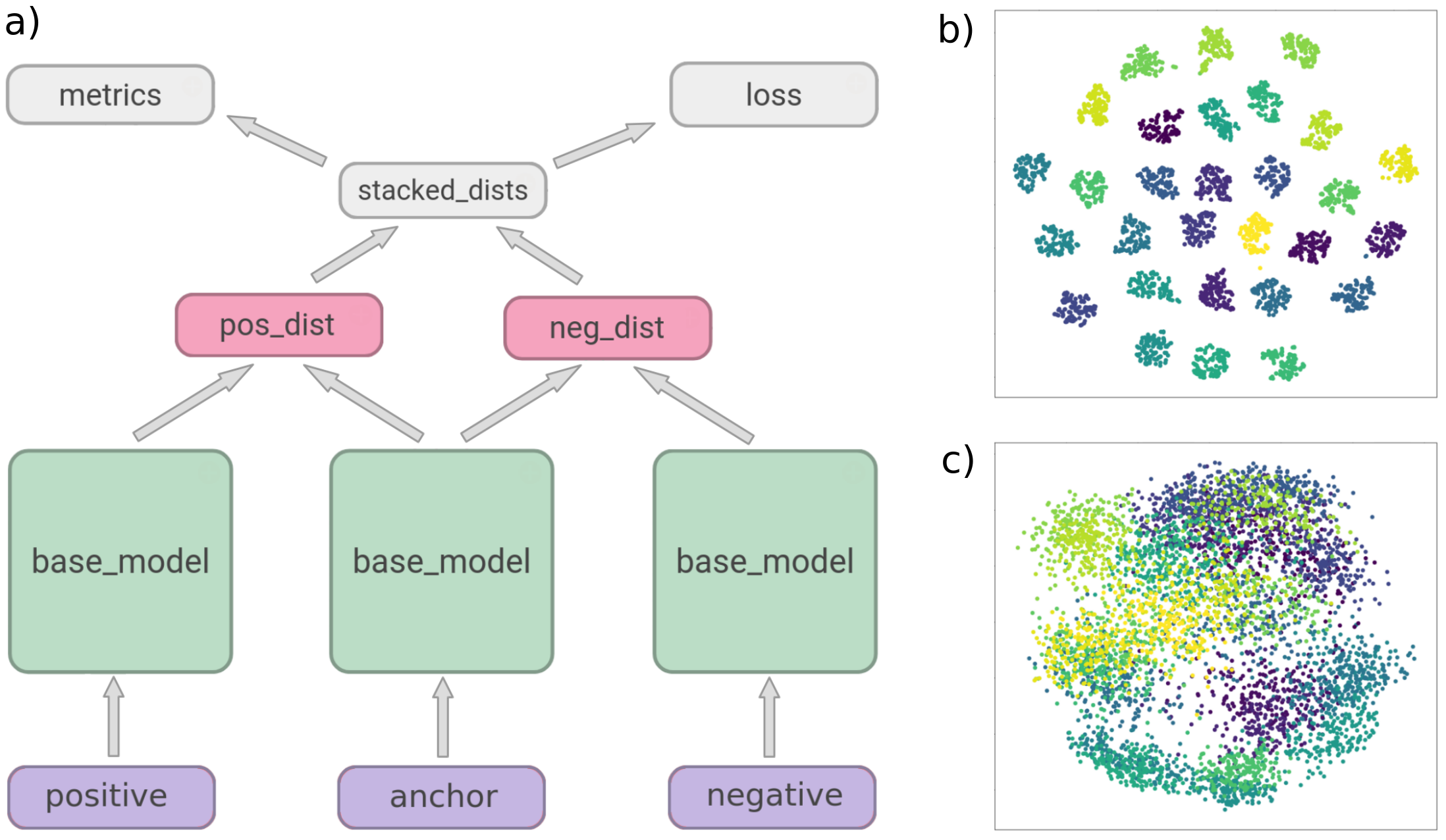}
\caption{\ref{fig:networks}a) Architecture of the Triplet network: three identical copies of the base network (VGG-16 or ResNet-50 followed by dimension-reducing, embedding-generating layers), followed by layers which compute embedding distances, loss and accuracy. \ref{fig:networks}b) t-SNE visualization of the embeddings generated by the Triplet network from the embeddings of the X-ray images of the 29 bones. \ref{fig:networks}c) Principle component analysis (PCA) visualization of the embeddings generated by the Triplet network from the embeddings of the X-ray images of the 29 bones.}
\label{fig:networks}
\end{figure}

The Triplet network architecture is shown in Figure \ref{fig:networks}.

\subsubsection*{Classification using bone fingerprints generated by the Triplet network}

After extracting embeddings for each of the 26100 bone images, a shallow classifier (a k-nearest neighbor classifier) was trained on these embeddings. The validation accuracy of this classifier was $100\%$, meaning that the classifier was able to assign these embeddings to their respective classes without any errors. This result was reproduced successfully with a support vector machine (SVM) \cite{SVM} classifier. It did not make a difference in accuracy whether the classifier was trained on embeddings from bones on which the neural network was also trained, or on completely new bones; we excluded five sample bones from the dataset, trained the Triplet network on the remaining 24 bones only, then generated embeddings for all the images of the five sample bones. Even after training the kNN classifier on the embeddings of the five excluded sample bones, the classifier still achieved a validation accuracy of $100\%$. This shows how well the Triplet network's ability to extract features characteristic to specific bones generalizes to previously unseen bones. Since the embeddings can be so accurately assigned to their classes (which represent specific bones), each embedding can be thought of as a de-facto unique fingerprint of a bone. 

\subsubsection*{Pairwise Euclidean distances between bone fingerprints}

After establishing that embeddings of different bones can be easily separated by a classifier, we examined under which conditions pairwise comparisons of images based on their embeddings can be performed. We conducted an experiment with the embeddings from images of the five bones the network had not been trained on; pairwise Euclidean distances were computed between all the images within the same class and between images of different classes. The results were then visualized in a scatter plot to show whether there was a distance threshold with all intra-class distances lying below it and all inter-class distances above it. This was not true of all the images. It does not, however, contradict the high kNN classifier accuracy, because classifying a group of images with a 100\% accuracy only requires the classifier to find the best possible match for the input data and does not pose the additional constraint of how far in the Euclidean space the embeddings of the non-matching class instances must lie.

A clear threshold separating \textit{all} intra-class embedding distances from \textit{all} inter-class distances was found after embeddings of outlier images were removed. Such outliers were images of bones which depicted the bone at an angle that deviated strongly from either the anteroposterior or the mediolateral view. The observed separating threshold coincided with the \textit{margin} value enforced during training, as was intuitively expected.

\begin{figure}[ht]
\centering
\includegraphics[width=\linewidth]{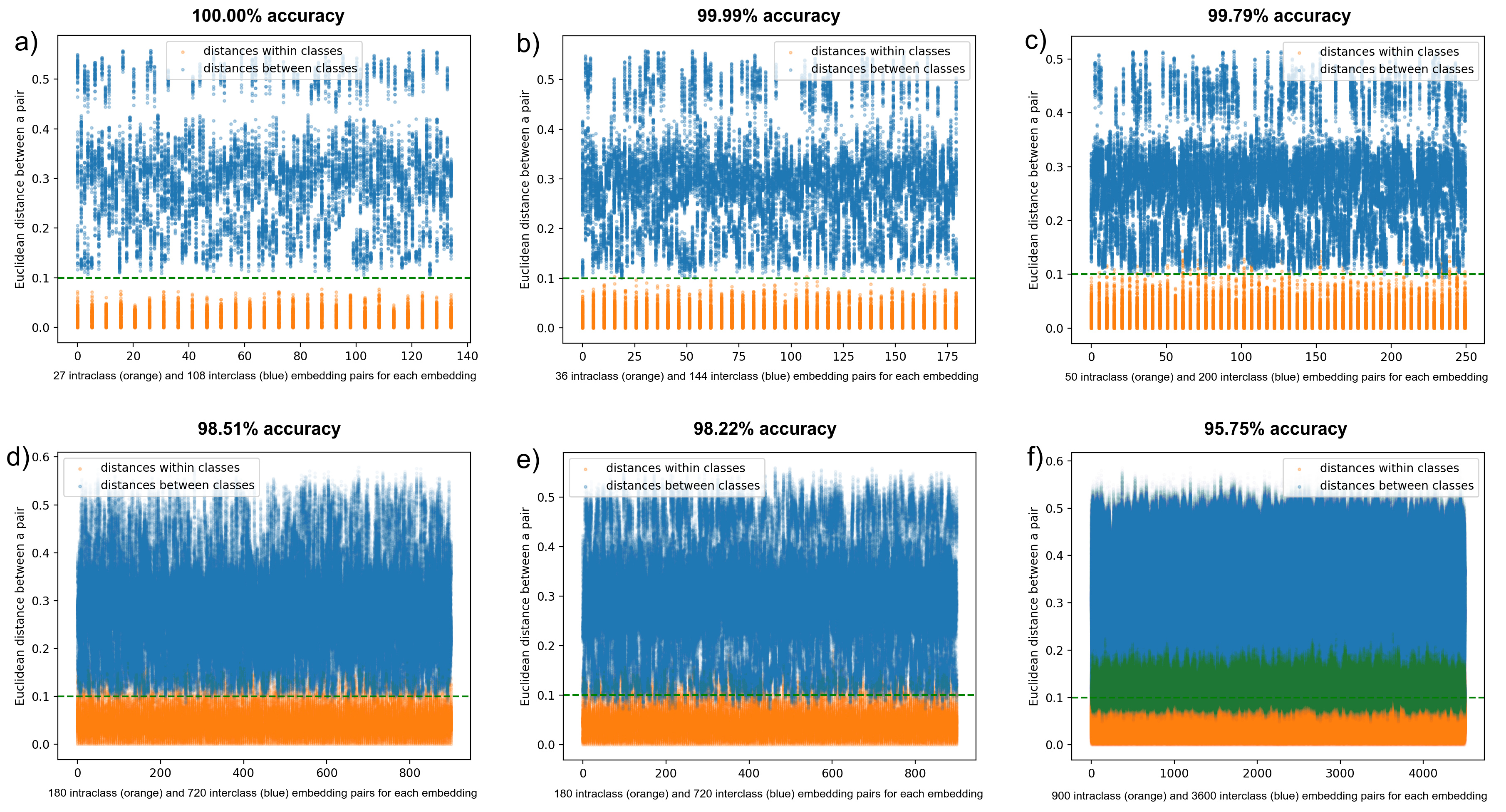} 
\caption{Distribution of pairwise Euclidean distances between embeddings of 4500 artificial X-ray images generated from five bones. The Triplet network was not trained on any images of these bones. \ref{fig:embedding-distances}a) Pairwise embedding distances between images depicting bones from an angle which did not deviate from the standard anteroposterior or mediolateral (AP/ML) view by more than four degrees in any direction. Radiation energy interval: 146-158 keV. 
\ref{fig:embedding-distances}b) Max. deviation from AP/ML view: four degrees. Radiation energy interval: 140-158 keV. 
\ref{fig:embedding-distances}c) Max. deviation from AP/ML view: seven degrees. Radiation energy interval: 140-158 keV. 
\ref{fig:embedding-distances}d) Max. deviation from AP/ML view: 22 degrees around the bone's longitudinal axis, four degrees around an axis perpendicular to image plane. Radiation energy interval: 140-158 keV. 
\ref{fig:embedding-distances}e) Max. deviation from AP/ML view: four degrees around the bone's longitudinal axis, 22 degrees around an axis perpendicular to image plane. Radiation energy interval: 140-158 keV. 
\ref{fig:embedding-distances}f) Max. deviation from AP/ML view: 22 degrees. Radiation energy interval: 140-158 keV. Overlapping dots are marked green.}
\label{fig:embedding-distances}
\end{figure}

Figure \ref{fig:embedding-distances} shows intra-class (orange) and inter-class (blue) embedding distances, separated by a gap in which the \textit{margin} lies after the outliers are removed. When only embeddings from images are used where the depicted bone deviates by no more than four degrees in either direction from the standard anteroposterior and mediolateral view and when the radiation energy used for the images is within the interval 146-158 keV, the \textit{margin} value separates the interclass and intraclass distances between embeddings with a 100.00\% accuracy. Angle deviations have a stronger impact on this distance accuracy; when embeddings are used which were generated from images encompassing the entire energy range used in generating the images (140-158 keV), the accuracy drops slightly to 99.99\%. If both angles deviate by no more than seven degrees, we achieve an accuracy of 99.79\%. If no more than one rotation angle deviates by no more than 22 degrees (and the second one by no more than four degrees), the accuracy is between 99.22\% and 99.51\%. If all generated images are used (angle deviations within a 22 degree interval, radiation energy interval 140-158 keV), the accuracy drops to 95.75\%.

\subsubsection*{Bone shape estimation}

We used the Triplet network based on VGG-16 in combination with a kNN classifier to estimate the 3D shape of bones, given their 2D X-ray images. The $99.9\%$ triplet accuracy and the $100\%$ kNN accuracy mean that this neural network and kNN combination is excellent at predicting exact matches for bones, given their 2D images.

For shape estimation of bones unfamiliar to the network, however, high-quality nearest matches are required. We tested how well our Triplet network/kNN classifier combination performs when forced to make a prediction for the input image of a bone which was not part of the classifier's dataset, i.e. when any match predicted by our classifier will not be an exact match, but rather a nearest match, based on the features extracted by the network. How good these nearest matches are was tested by classifying image embeddings of the five sample bones excluded from the dataset the neural network was trained on, using the kNN classifier.

\subsubsection*{Evaluation}

The evaluation of how well the predicted 3D shapes matched the true 3D shapes of the bones presented to the network as artificial 2D X-ray images was performed by computing the RMS distance and the Hausdorff distance between the predicted 3D shapes of each of the five sample bones and their true 3D shapes. The result was that on average, no more than $2.6$ of the 24 bones would have been better matches. I.e. only $10.8\%$ of the bones the network had been trained on would have matched the shape of the bones in the input images better than the bones chosen by the classifier. In the best case, the best available match was predicted by the classifier. In the worst case, seven bones other than the one predicted would have been better matches. In the remaining three cases, one, two and three bones would have been better matches.

\subsubsection*{Evaluation by computing distances between predicted and true 3D shapes}

How good a 3D shape match was and also how many other candidates would have been better matches was evaluated by computing the RMS and Hausdorff distance between the predicted and the true 3D shape. For this purpose, the surface meshes of the bone from the input image and the predicted bone were aligned using the \textit{MeshLab} \cite{meshlab} software and the RMS and Hausdorff distance between them was computed (also in \textit{MeshLab}), both in millimeters and with respect to the bounding box diagonal. Table \ref{tab:RMS-Hausdorff} summarizes the results.

\begin{table}
    \centering
    \begin{tabular}{|l | l | l | l | l|}
    \hline
    Sample bone & Absolute RMS dist. & Absolute Hausdorff dist. & Relative RMS dist. & Relative Hausdorff dist. \\
    \hline
    \hline
    S1 & 0.77 mm & 2.23 mm & 0.0065 & 0.0188 \\ 
    \hline
    S2 & 0.72 mm & 2.89 mm & 0.0066 & 0.0266 \\
    \hline
    S3 & 0.73 mm & 2.72 mm & 0.0058 & 0.0217 \\
    \hline
    S4 & 1.26 mm & 5.09 mm & 0.0114 & 0.0462 \\
    \hline
    S5 & 0.82 mm & 2.62 mm & 0.0061 & 0.0195 \\
    \hline
    S1-S5 average & 0.86 mm & 3.11 mm & 0.0073 & 0.0266 \\
    \hline
    \end{tabular}
    \caption{Root Mean Square (RMS) distances and Hausdorff distances between the five sample bones S1-S5 and their matches predicted by the VGG-based Triplet network and kNN classifier on the 2D image dataset of bones scaled according to their real-world proportions. Absolute RMS distance is given in $mm$, relative distances are given with respect to the bounding box diagonal}
    \label{tab:RMS-Hausdorff}
\end{table}

\begin{figure}[ht]
\centering
\includegraphics[width=\linewidth]{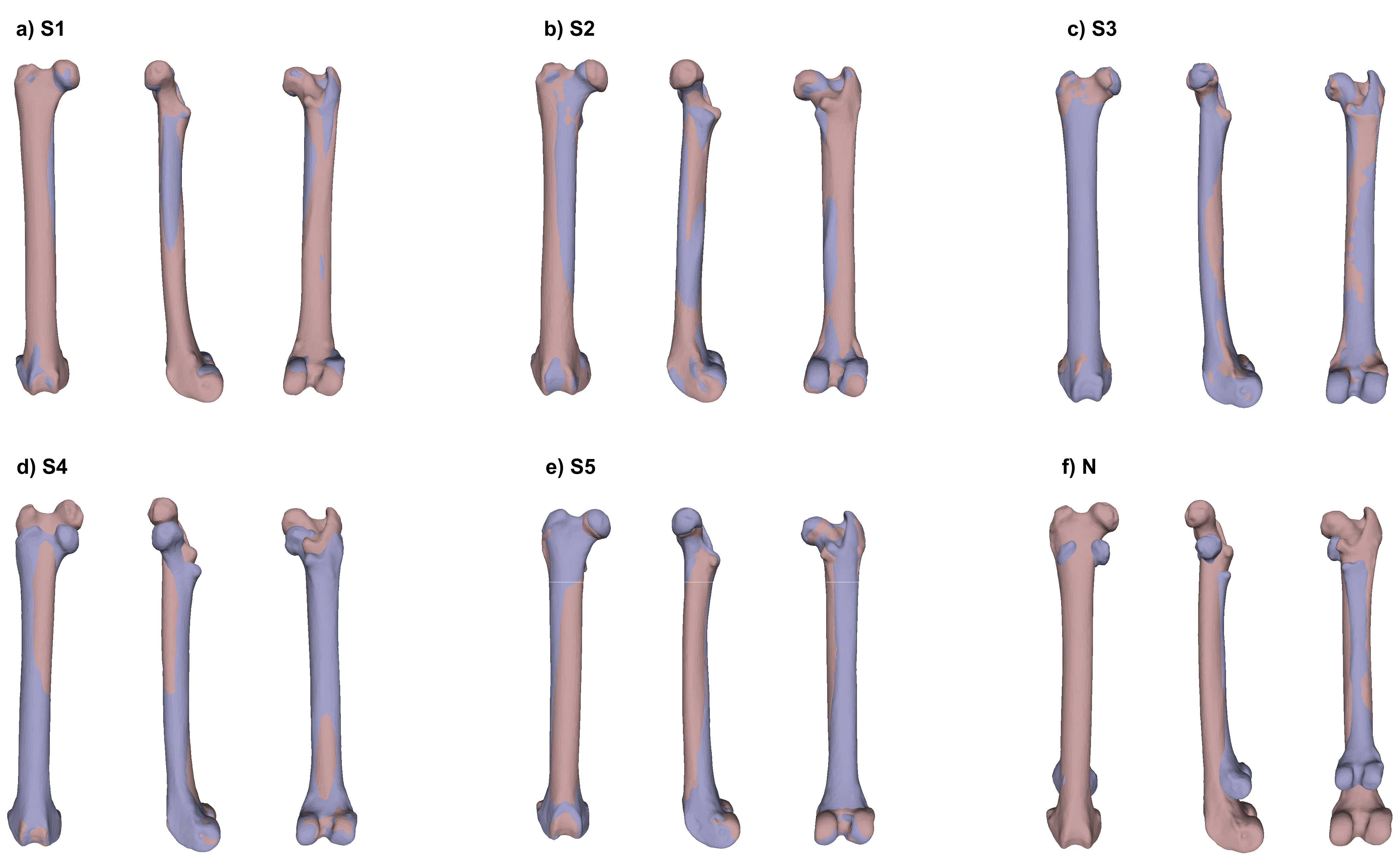}
\caption{\ref{fig:alignments}a)-\ref{fig:alignments}e) Alignments between the five sample bones \textbf{S1-S5} and their closest matches as predicted by the VGG-based Triplet network in combination with the kNN classifier. \ref{fig:alignments}f) Alignment between the pair of bones which are the worst possible match present in the dataset, based on their RMS distance.}
\label{fig:alignments}
\end{figure}

Figure \ref{fig:alignments} shows the results of the qualitative evaluation - alignments between the 3D shapes predicted by the network-kNN combination and the true 3D shapes. For comparison, an alignment between two bones which are the worst possible pairwise match in the dataset is also shown.

\subsubsection*{Evaluation by comparing to statistical shape models and other approaches}

We also evaluated the quality of our 3D matches by comparing them with match quality achieved by a statistical shape models implementation. For this purpose, we used three additional 3D CT scans that were not part of the original dataset of 29 bones. For each of the new bones, a 3D DICOM file as well as two corresponding natural X-ray scans were available. The first natural X-ray scan showed the bone from the anteroposterior and the second from the mediolateral view. We let the network predict the closest 3D match for the natural X-ray scans out of the original 24 bones. We then used the SSM software to generate a statistical shape model from those 24 bones. The SSM model based on the 24 bones was then deformed by fitting it to the bones shown in the natural 2D X-ray scans. The result of the fitting process was exported as a surface model and compared with the shape match predicted by our network/kNN classifier.

In two out of three cases, our network combined with the kNN lead to a better match. In the third case, the SSM produced a better match. The RMS distances between the predicted and true shapes when using our method were $0.89mm$, $2.53mm$, and $0.93mm$. The RMS distances when using the SSM method were $1.35mm$, $3.07mm$, and $0.80mm$, respectively. It is notable that in the case of the second bone, where our model achieved an unusually high RMS error, and also in the case of the third bone, where our model performed worse than the SSM approach, the bone in the natural X-ray image was approx. $6\%$ shorter in the anteroposterior view than in the lateral view, an issue which occurs in practice due to difficulties with positioning the patient in the CT machine. The network was, however, not trained on bones with different lengths in the two views.

A comparison with other approaches to 2D to 3D bone shape estimation found through literature research shows that our approach can compete with them; for the eight bones we examined (five sample bones and three bones paired with their natural 2D X-ray images), the average RMS distance between them and their match predicted by the Triplet network/kNN combination was $1.08mm$ ($0.89mm$ for the five sample bones where predictions were made based on artificial X-ray images and $1.45mm$ for the three bones where predictions were made from natural X-ray images).

In comparison, the average mean or RMS distance between predicted and true shapes achieved by eight other examined approaches was $1.32mm$. These approaches include: statistical shape models, Laplacian surface deformation (LSD), a fast Fourier transform (FFD)-based method, a method based on iterative nonrigid 2D point-matching process and thin-plate spline-based deformation, and non-stereo corresponding contour (NSCC) method \cite{3Dlaplace_peter}.


\section*{Discussion}


In this paper, we presented multiple ways in which deep neural networks are capable of working with X-ray images of bones. Both transfer learning and triplet loss training in combination with a kNN classifier showed that neural networks pre-trained on the ImageNet dataset and fine-tuned on a dataset of artificially generated bone X-ray images are able to differentiate between femurs from different cat specimens with an accuracy of $100.0\%$. This result demonstrates neural networks' ability to extract meaningful features from X-ray images and opens up new possibilities for deep learning-based, fully automated work with X-ray images of bones.

We tested our triplet loss-trained neural network (combined with a kNN classifier) on images of eight bones to determine their 3D shape by selecting the best-fitting matches out of a set of 24 choices. Our network first extracted a 128-dimensional embedding from each input image, and a kNN-classifier then determined the best match based on embedding distances in the Euclidean space. The average root mean squared distance between our shape predictions and the ground truth was $1.08mm$.

The comparison with existing 2D to 3D bone shape estimation approaches shows that our approach can compete with them and even seems to perform better than the average achieved by eight other approaches we examined \cite{3Dlaplace_peter}, even though an exact comparison is not possible, since the respective publications use absolute shape distances in \textit{mm}, not relative ones (normalized by their bounding box diagonal). 

A clear advantage in comparison with other methods such as statistical shape models is that our approach is completely domain-agnostic and as such does not require any previous knowledge about the bones in our dataset, such as their geometry. Re-training our classifier to use completely different bones is as easy as supplying it with a new set of 3D CT scans. After a training dataset of images has been generated, our network learns to extract all features necessary for bone classification and shape estimation automatically.

A very important finding is that a neural network trained using the triplet loss method is able to determine the identity of a bone based only on its 2D X-ray image. The 128-dimensional embedding our network generates for each input image can either be used for shape estimation by letting a kNN classifier determine the nearest match for an unknown bone, or for determining the identity of a bone. Determining bone identity is possible either by using a kNN classifier trained on embeddings of bones that are possible matches for the bone in question, or by pairwise comparisons directly, using the triplet \textit{margin} as the minimum Euclidean distance between embeddings of the bone images being compared. When using pairwise comparisons, we achieved a 96\% accuracy for a total of 4500 X-ray images generated from 3D CT scans of five bones the Triplet network had not been trained on. The 96\% accuracy was achieved when the rotation angle varied within a 22 degree interval. A 100\% accuracy was achieved when the rotation angle was restricted to an 8 degree interval. 

One possible application for the 3D shape estimation described here is manufacturing certain types of casts for patients whose bones could not undergo a 3D imaging procedure directly, either due to technical availability or health issues.

In addition, our findings about bone fingerprinting have interesting implications for database systems - such fingerprints may be used to either enrich bone databases by highly compressed information about the bones, or they can even be used as a search key. This has possible forensic applications where image-based look-up of bones with unknown origin in bone databases would be sped up significantly and could be performed within seconds.


\section*{Methods}

\subsection*{Dataset}

\textbf{2D and 3D data.} Our dataset consisted of 29 3D CT scans of femurs of 29 different cats in the DICOM format. Using the software \textit{MeVisLab}, we generated 900 artificial 2D X-ray images for each of the 29 CT scans. For this purpose, we created a \textit{MeVisLab} macromodule which contained a Python script. In this script, we set the values for rotation around the X axis to be within the interval [70, 112] and sampled every third value from this interval: 70, 73, 76..., 112 . For the rotation around the Y axis, the interval was set to [-21, 22] and we also sampled every third value. For the radiation energy, we sampled every sixth value from the interval [140, 161]. We sampled these values in a triple nested loop for each DICOM input file, which resulted in $15 \cdot 15 \cdot 4 = 900$ different value combinations. The rotations around the two axes rotated the bone slightly around its own longitudinal axis as well as around an axis perpendicular to the 2D image plane. In the innermost of the nested loops, the sampled values were assigned to the \textit{DRR} and \textit{DRRLUT} module fields of our \textit{MeVisLab} network, after which the resulting image was saved in the PNG format using the \textit{ImageSave} Module. The resulting images were slightly blurred using a sigma value of $1.0$ to remove minor CT artifacts \cite{CT_artifacts}. This way, we automatically generated 26100 images from the 29 DICOM files. These 26100 images served as the dataset to train and evaluate our neural network/kNN classifier on.

\textbf{Standard viewing angles}. For each DICOM file, we used \textit{OrthoSwapFlip MeVisLab} module to rotate the bone by 90 degrees. After this, we had 29 DICOMs showing a bone from the anteroposterior and 29 from the mediolateral view. This way, we could more easily generate 900 400x800 pixel images with the anteroposterior and 900 400x800 pixel images with the mediolateral view. We then merged these two groups, creating a dataset of 900 800x800 pixel images per class, each showing the same bone from the two orthogonal angles next to each other. This resolution is higher than many convolutional neural networks accept; this was on purpose so that no upscaling would be needed for the networks.

\textbf{Bone image scaling}. We also created a version of our 2D image dataset which contained images of bones scaled according to their real-world proportions. This dataset version achieved slightly higher Triplet network accuracy, and a much higher accuracy when 3D shapes were predicted. The bones depicted in the images were scaled, without modifying the image pixel resolution, the following way: A scaling object with a known real-world size (10 mm) was drawn into all DICOM volumes using the \textit{MeVisLab} module \textit{DrawVoxels3D}. 2D images were then generated from these modified DICOM files, one image from the anteroposterior DICOM file of a bone, the second image from the corresponding mediolateral DICOM file of the same bone. The reason was that the DICOM volumes were not cubic, nor was the bone always positioned perfectly in the middle, which is why the anteroposterior and mediolateral  DICOM volumes had to be scaled separately. The pixel sizes of the scaling objects next to all bones were measured and scaling coefficients were calculated to determine by how many percent the content of each image had to be shrunk to make the scaling objects in all images have the same pixel size. The images of the entire 2D image dataset were then shrunk according to these coefficients. Anteroposterior and mediolateral views of the same image could have different coefficients and had to be shrunk separately. After shrinking, each image was padded with black borders (the background color) to bring their resolution back to 400x800 pixels.

\textbf{Surface model extraction for evaluation}. For evaluation purposes, the surface model (mesh) of each 3D CT scan was extracted so that the difference between a predicted shape and the true shape could be measured. First, a mesh in the STL format was extracted from each DICOM file using the \textit{MeVisLab} modules \textit{WEMIsoSurface} and \textit{WEMSave}. Setting the Iso value in \textit{WEMIsoSurface} to 1370 led to successful removal of the soft tissue of the DICOM files, so that only the bone tissue remained and could be exported as a mesh. This mesh was then post-processed to remove the inner bone structure as well as the neighboring bones (cat pelvis, tibia and patella). The post-processing was done in the \textit{Materialise Mimics} software. A viable alternative is using the free software \textit{MeshLab} and the freely available \textit{Microsoft 3D Builder}. In \textit{MeshLab}, the bone artifacts can be selected both manually (for removing the neighboring bones) and by the degree of their occlusion (for removing the inner bone parts). Closing the holes caused by removal of neighboring bones can be done when importing the meshes into \textit{Microsoft 3D Builder} - the software merely asks whether any mesh holes should be closed before importing. However, \textit{Materialise Mimics} lead to better results than the free alternatives.

\textbf{Natural X-ray scan pre-processing}. The natural X-ray scans used for comparison to SSM were pre-processed by cropping them approximately the same way the artificial X-ray scans were cropped - closely enough that only a minimum of the neighboring bones was visible. These X-ray scans were made with a scaling object of known real-world size present on the X-ray table, so that they could be scaled in proportion to the artificial X-ray scans on which the network had been trained.

\subsection*{Neural Network Architecture}

The triplet loss training was implemented by building all the necessary computations directly into the neural network architecture. We used first the neural network ResNet-50 \cite{resnet} and then the network VGG-16 \cite{vgg} as the base building blocks of the Triplet network.

\textbf{ResNet-50} \cite{resnet} is a 50 layer deep convolutional neural network first published by in 2015 by He et al. It achieves a $6.71\%$ top-5 error rate on the benchmark ImageNet \cite{imagenet} dataset. Each layer in the network extracts a new layer of features, with a classifier at the end. The number of layers has been shown to be of crucial importance. However, stacking too many layers, even if the problem of exploding/vanishing gradients is taken care of, leads to a performance degradation. He et al. solve this problem by making heavy use of so-called shortcut (residual) connections between the layers of their ResNet networks. This way, networks of over 150 layers have been successfully trained. We used the 50-layer deep variant of the ResNet architecture as basis for our feature extractor within the Triplet network, on the assumption that the features present in the X-ray images might be complex enough and thus require such a large number of abstraction layers.

\textbf{VGG-16} \cite{vgg} is a 16 layer deep convolutional network published by Simonyan and Zisserman in 2014. It achieves a $9.33\%$ top-5-error rate on the benchmark ImageNet dataset. What distinguishes it from its predecessors is that it successfully trades larger convolutional filters for smaller ones stacked behind one another along the length of the network, effectively covering a larger receptive field using a smaller number of network parameters. Similarly to ResNet, it also performed a significant increase of the number of its layers when compared to its predecessors. However, with 16 layers, it is much shorter than ResNet-50, and its architecture is also different. We used VGG-16 in later experiments to cover the case that the features in the X-ray images might be too simple for ResNet-50 and a more shallow network might extract features better suited for shape estimation. Our experiments showed that this was, in fact, the case. While the triplet accuracy was only slightly improved by using VGG-16 instead of ResNet-50 (by $0.1\%$), the 3D shape predictions were improved by $4.2\%$. Using ResNet-50, an average of 3.6 out of 24 bones would have been better shape matches for out five sample bones. When using VGG-16, 2.6 of the bones not predicted by our classifier would have been better matches.

\textbf{Triplet network.} The Triplet network used to generate embeddings of the input X-ray images was built the following way: It consisted of three identical copies of the so-called base network. ResNet-50, in later experiments VGG-16, in combination with dimension-reducing layers, was used as the base network. Each of the three identical base network copies received a different triplet image as input data: One base network received the so-called \textit{anchor} image (which was randomly selected from the training set), the second base network received the \textit{positive} image (image from the same class as \textit{anchor}), and the third one received the \textit{negative} image (image from a different class than \textit{anchor}). The base networks themselves had to be modified, because they had to be used in a capacity as feature extractors, not classifiers. To achieve this, we removed their last (softmax classifier) layer and kept the networks only up to their last fully-connected layer, which produced an embedding with over 1000 dimensions. We then reduced this output down to $d$ dimensions, with $d$ set to 128, 64, 32 and 16 during our experiments. We performed this dimension reduction by appending three pairs of layers - a dimension-reducing fully-connected layer followed by an L2 regularization layer. The last of the three fully-connected layers produced the $d$-dimensional output, and the last L2 layer normalized it to have unit length. The layer that followed combined the embeddings by stacking them together so that they would be available for the accuracy and loss functions.

The triplet loss was defined to be zero when the difference between the Euclidean distance of \textit{anchor} and \textit{positive} embedding and the Euclidean distance of \textit{anchor} and \textit{negative} embedding was larger than \textit{margin}, as shown in equation \ref{eq:triplet_accuracy}.


\bibliography{main} 

\begin{thebibliography}{10}
\urlstyle{rm}
\expandafter\ifx\csname url\endcsname\relax
  \def\url#1{\texttt{#1}}\fi
\expandafter\ifx\csname urlprefix\endcsname\relax\def\urlprefix{URL }\fi
\expandafter\ifx\csname doiprefix\endcsname\relax\def\doiprefix{DOI: }\fi
\providecommand{\bibinfo}[2]{#2}
\providecommand{\eprint}[2][]{\url{#2}}

\bibitem{3Dlaplace_peter}
\bibinfo{author}{Karade, V.} \& \bibinfo{author}{Ravi, B.}
\newblock \bibinfo{journal}{\bibinfo{title}{3d femur model reconstruction from
  biplane x-ray images: a novel method based on laplacian surface
  deformation}}.
\newblock {\emph{\JournalTitle{International journal of computer assisted
  radiology and surgery}}} \textbf{\bibinfo{volume}{10}},
  \bibinfo{pages}{473--485} (\bibinfo{year}{2015}).

\bibitem{SSM_baka}
\bibinfo{author}{Baka, N.} \emph{et~al.}
\newblock \bibinfo{journal}{\bibinfo{title}{2d--3d shape reconstruction of the
  distal femur from stereo x-ray imaging using statistical shape models}}.
\newblock {\emph{\JournalTitle{Medical image analysis}}}
  \textbf{\bibinfo{volume}{15}}, \bibinfo{pages}{840--850}
  (\bibinfo{year}{2011}).

\bibitem{FFD_filippi}
\bibinfo{author}{Filippi, S.}, \bibinfo{author}{Motyl, B.} \&
  \bibinfo{author}{Bandera, C.}
\newblock \bibinfo{journal}{\bibinfo{title}{Analysis of existing methods for 3d
  modelling of femurs starting from two orthogonal images and development of a
  script for a commercial software package}}.
\newblock {\emph{\JournalTitle{Computer methods and programs in biomedicine}}}
  \textbf{\bibinfo{volume}{89}}, \bibinfo{pages}{76--82}
  (\bibinfo{year}{2008}).

\bibitem{SSM_fleute}
\bibinfo{author}{Fleute, M.} \& \bibinfo{author}{Lavall{\'e}e, S.}
\newblock \bibinfo{title}{Nonrigid 3-d/2-d registration of images using
  statistical models}.
\newblock In \emph{\bibinfo{booktitle}{International Conference on Medical
  Image Computing and Computer-Assisted Intervention}},
  \bibinfo{pages}{138--147} (\bibinfo{organization}{Springer},
  \bibinfo{year}{1999}).

\bibitem{nonSSM_gamage}
\bibinfo{author}{Gamage, P.}, \bibinfo{author}{Xie, S.~Q.},
  \bibinfo{author}{Delmas, P.} \& \bibinfo{author}{Xu, P.}
\newblock \bibinfo{title}{3d reconstruction of patient specific bone models
  from 2d radiographs for image guided orthopedic surgery}.
\newblock In \emph{\bibinfo{booktitle}{Digital Image Computing: Techniques and
  Applications, 2009. DICTA'09.}}, \bibinfo{pages}{212--216}
  (\bibinfo{organization}{IEEE}, \bibinfo{year}{2009}).

\bibitem{nonSSM_laport}
\bibinfo{author}{Laporte, S.}, \bibinfo{author}{Skalli, W.},
  \bibinfo{author}{De~Guise, J.}, \bibinfo{author}{Lavaste, F.} \&
  \bibinfo{author}{Mitton, D.}
\newblock \bibinfo{journal}{\bibinfo{title}{A biplanar reconstruction method
  based on 2d and 3d contours: application to the distal femur}}.
\newblock {\emph{\JournalTitle{Computer Methods in Biomechanics \& Biomedical
  Engineering}}} \textbf{\bibinfo{volume}{6}}, \bibinfo{pages}{1--6}
  (\bibinfo{year}{2003}).

\bibitem{nonSSM_tang}
\bibinfo{author}{Tang, T.} \& \bibinfo{author}{Ellis, R.}
\newblock \bibinfo{title}{2d/3d deformable registration using a hybrid atlas}.
\newblock In \emph{\bibinfo{booktitle}{International Conference on Medical
  Image Computing and Computer-Assisted Intervention}},
  \bibinfo{pages}{223--230} (\bibinfo{organization}{Springer},
  \bibinfo{year}{2005}).

\bibitem{SSM_zhu}
\bibinfo{author}{Zhu, Z.} \& \bibinfo{author}{Li, G.}
\newblock \bibinfo{journal}{\bibinfo{title}{Construction of 3d human distal
  femoral surface models using a 3d statistical deformable model}}.
\newblock {\emph{\JournalTitle{Journal of biomechanics}}}
  \textbf{\bibinfo{volume}{44}}, \bibinfo{pages}{2362--2368}
  (\bibinfo{year}{2011}).

\bibitem{galibarov}
\bibinfo{author}{Galibarov, P.}, \bibinfo{author}{Prendergast, P.} \&
  \bibinfo{author}{Lennon, A.}
\newblock \bibinfo{journal}{\bibinfo{title}{A method to reconstruct
  patient-specific proximal femur surface models from planar pre-operative
  radiographs}}.
\newblock {\emph{\JournalTitle{Medical engineering \& physics}}}
  \textbf{\bibinfo{volume}{32}}, \bibinfo{pages}{1180--1188}
  (\bibinfo{year}{2010}).

\bibitem{prakoonwit}
\bibinfo{author}{Prakoonwit, S.}
\newblock \bibinfo{journal}{\bibinfo{title}{Towards multiple 3d bone surface
  identification and reconstruction using few 2d x-ray images for
  intraoperative applications}}.
\newblock {\emph{\JournalTitle{International Journal of Art, Culture and Design
  Technologies (IJACDT)}}} \textbf{\bibinfo{volume}{4}},
  \bibinfo{pages}{13--31} (\bibinfo{year}{2014}).

\bibitem{triplet_network}
\bibinfo{author}{Hoffer, E.} \& \bibinfo{author}{Ailon, N.}
\newblock \bibinfo{title}{Deep metric learning using triplet network}.
\newblock In \emph{\bibinfo{booktitle}{International Workshop on
  Similarity-Based Pattern Recognition}}, \bibinfo{pages}{84--92}
  (\bibinfo{organization}{Springer}, \bibinfo{year}{2015}).

\bibitem{facenet}
\bibinfo{author}{Schroff, F.}, \bibinfo{author}{Kalenichenko, D.} \&
  \bibinfo{author}{Philbin, J.}
\newblock \bibinfo{title}{Facenet: A unified embedding for face recognition and
  clustering}.
\newblock In \emph{\bibinfo{booktitle}{Proceedings of the IEEE conference on
  computer vision and pattern recognition}}, \bibinfo{pages}{815--823}
  (\bibinfo{year}{2015}).

\bibitem{person_reident}
\bibinfo{author}{Hermans, A.}, \bibinfo{author}{Beyer, L.} \&
  \bibinfo{author}{Leibe, B.}
\newblock \bibinfo{journal}{\bibinfo{title}{In defense of the triplet loss for
  person re-identification}}.
\newblock {\emph{\JournalTitle{CoRR}}}
  \textbf{\bibinfo{volume}{abs/1703.07737}} (\bibinfo{year}{2017}).

\bibitem{triplet_Cheng_2016_CVPR}
\bibinfo{author}{Cheng, D.}, \bibinfo{author}{Gong, Y.}, \bibinfo{author}{Zhou,
  S.}, \bibinfo{author}{Wang, J.} \& \bibinfo{author}{Zheng, N.}
\newblock \bibinfo{title}{Person re-identification by multi-channel parts-based
  cnn with improved triplet loss function}.
\newblock In \emph{\bibinfo{booktitle}{The IEEE Conference on Computer Vision
  and Pattern Recognition (CVPR)}} (\bibinfo{year}{2016}).

\bibitem{triplet_G_2016_CVPR}
\bibinfo{author}{Kumar B~G, V.}, \bibinfo{author}{Carneiro, G.} \&
  \bibinfo{author}{Reid, I.}
\newblock \bibinfo{title}{Learning local image descriptors with deep siamese
  and triplet convolutional networks by minimising global loss functions}.
\newblock In \emph{\bibinfo{booktitle}{The IEEE Conference on Computer Vision
  and Pattern Recognition (CVPR)}} (\bibinfo{year}{2016}).

\bibitem{triplet_Zhuang_2016_CVPR}
\bibinfo{author}{Zhuang, B.}, \bibinfo{author}{Lin, G.}, \bibinfo{author}{Shen,
  C.} \& \bibinfo{author}{Reid, I.}
\newblock \bibinfo{title}{Fast training of triplet-based deep binary embedding
  networks}.
\newblock In \emph{\bibinfo{booktitle}{The IEEE Conference on Computer Vision
  and Pattern Recognition (CVPR)}} (\bibinfo{year}{2016}).

\bibitem{triplet_He_2018_CVPR}
\bibinfo{author}{He, X.}, \bibinfo{author}{Zhou, Y.}, \bibinfo{author}{Zhou,
  Z.}, \bibinfo{author}{Bai, S.} \& \bibinfo{author}{Bai, X.}
\newblock \bibinfo{title}{Triplet-center loss for multi-view 3d object
  retrieval}.
\newblock In \emph{\bibinfo{booktitle}{The IEEE Conference on Computer Vision
  and Pattern Recognition (CVPR)}} (\bibinfo{year}{2018}).

\bibitem{triplet_bui2017compact}
\bibinfo{author}{Bui, T.}, \bibinfo{author}{Ribeiro, L.},
  \bibinfo{author}{Ponti, M.} \& \bibinfo{author}{Collomosse, J.}
\newblock \bibinfo{journal}{\bibinfo{title}{Compact descriptors for
  sketch-based image retrieval using a triplet loss convolutional neural
  network}}.
\newblock {\emph{\JournalTitle{Computer Vision and Image Understanding}}}
  \textbf{\bibinfo{volume}{164}}, \bibinfo{pages}{27--37}
  (\bibinfo{year}{2017}).

\bibitem{triplet_liu2016multi}
\bibinfo{author}{Liu, J.} \emph{et~al.}
\newblock \bibinfo{title}{Multi-scale triplet cnn for person
  re-identification}.
\newblock In \emph{\bibinfo{booktitle}{Proceedings of the 24th ACM
  international conference on Multimedia}}, \bibinfo{pages}{192--196}
  (\bibinfo{organization}{ACM}, \bibinfo{year}{2016}).

\bibitem{metro}
\bibinfo{author}{Cignoni, P.}, \bibinfo{author}{Rocchini, C.} \&
  \bibinfo{author}{Scopigno, R.}
\newblock \bibinfo{title}{Metro: measuring error on simplified surfaces}.
\newblock In \emph{\bibinfo{booktitle}{Computer Graphics Forum}},
  vol.~\bibinfo{volume}{17}, \bibinfo{pages}{167--174}
  (\bibinfo{organization}{Wiley Online Library}, \bibinfo{year}{1998}).

\bibitem{hausdorff_dist_paper}
\bibinfo{author}{Aspert, N.}, \bibinfo{author}{Santa-Cruz, D.} \&
  \bibinfo{author}{Ebrahimi, T.}
\newblock \bibinfo{title}{Mesh: Measuring errors between surfaces using the
  hausdorff distance}.
\newblock In \emph{\bibinfo{booktitle}{Multimedia and Expo, 2002. ICME'02.
  Proceedings. 2002 IEEE International Conference on}},
  vol.~\bibinfo{volume}{1}, \bibinfo{pages}{705--708}
  (\bibinfo{organization}{IEEE}, \bibinfo{year}{2002}).

\bibitem{xray_positioning_variability}
\bibinfo{author}{Dewi, D. E.~O.} \emph{et~al.}
\newblock \bibinfo{journal}{\bibinfo{title}{Reproducibility of standing posture
  for x-ray radiography: a feasibility study of the balancaid with healthy
  young subjects}}.
\newblock {\emph{\JournalTitle{Annals of biomedical engineering}}}
  \textbf{\bibinfo{volume}{38}}, \bibinfo{pages}{3237--3245}
  (\bibinfo{year}{2010}).

\bibitem{radiation_body_weight}
\bibinfo{author}{Ching, W.}, \bibinfo{author}{Robinson, J.} \&
  \bibinfo{author}{McEntee, M.}
\newblock \bibinfo{journal}{\bibinfo{title}{Patient-based radiographic exposure
  factor selection: a systematic review}}.
\newblock {\emph{\JournalTitle{Journal of medical radiation sciences}}}
  \textbf{\bibinfo{volume}{61}}, \bibinfo{pages}{176--190}
  (\bibinfo{year}{2014}).

\bibitem{midlevel_transfer}
\bibinfo{author}{Oquab, M.}, \bibinfo{author}{Bottou, L.},
  \bibinfo{author}{Laptev, I.} \& \bibinfo{author}{Sivic, J.}
\newblock \bibinfo{title}{Learning and transferring mid-level image
  representations using convolutional neural networks}.
\newblock In \emph{\bibinfo{booktitle}{Computer Vision and Pattern Recognition
  (CVPR), 2014 IEEE Conference on}}, \bibinfo{pages}{1717--1724}
  (\bibinfo{organization}{IEEE}, \bibinfo{year}{2014}).

\bibitem{imagenet}
\bibinfo{author}{Deng, J.} \emph{et~al.}
\newblock \bibinfo{title}{{ImageNet: A Large-Scale Hierarchical Image
  Database}}.
\newblock In \emph{\bibinfo{booktitle}{CVPR09}} (\bibinfo{year}{2009}).

\bibitem{transfer_learning_softmax}
\bibinfo{author}{Huang, J.-T.}, \bibinfo{author}{Li, J.}, \bibinfo{author}{Yu,
  D.}, \bibinfo{author}{Deng, L.} \& \bibinfo{author}{Gong, Y.}
\newblock \bibinfo{title}{Cross-language knowledge transfer using multilingual
  deep neural network with shared hidden layers}.
\newblock In \emph{\bibinfo{booktitle}{Acoustics, Speech and Signal Processing
  (ICASSP), 2013 IEEE International Conference on}},
  \bibinfo{pages}{7304--7308} (\bibinfo{organization}{IEEE},
  \bibinfo{year}{2013}).

\bibitem{resnet}
\bibinfo{author}{He, K.}, \bibinfo{author}{Zhang, X.}, \bibinfo{author}{Ren,
  S.} \& \bibinfo{author}{Sun, J.}
\newblock \bibinfo{title}{Deep residual learning for image recognition}.
\newblock In \emph{\bibinfo{booktitle}{Proceedings of the IEEE conference on
  computer vision and pattern recognition}}, \bibinfo{pages}{770--778}
  (\bibinfo{year}{2016}).

\bibitem{vgg}
\bibinfo{author}{Simonyan, K.} \& \bibinfo{author}{Zisserman, A.}
\newblock \bibinfo{journal}{\bibinfo{title}{Very deep convolutional networks
  for large-scale image recognition}}.
\newblock {\emph{\JournalTitle{CoRR}}} \textbf{\bibinfo{volume}{abs/1409.1556}}
  (\bibinfo{year}{2014}).

\bibitem{gradCAM}
\bibinfo{author}{Selvaraju, R.~R.} \emph{et~al.}
\newblock \bibinfo{title}{Grad-cam: Visual explanations from deep networks via
  gradient-based localization}.
\newblock In \emph{\bibinfo{booktitle}{Proceedings of the IEEE International
  Conference on Computer Vision}}, \bibinfo{pages}{618--626}
  (\bibinfo{year}{2017}).

\bibitem{audioset_paper}
\bibinfo{author}{Hershey, S.} \emph{et~al.}
\newblock \bibinfo{title}{Cnn architectures for large-scale audio
  classification}.
\newblock In \emph{\bibinfo{booktitle}{2017 ieee international conference on
  acoustics, speech and signal processing (icassp)}}, \bibinfo{pages}{131--135}
  (\bibinfo{organization}{IEEE}, \bibinfo{year}{2017}).

\bibitem{SVM}
\bibinfo{author}{Cortes, C.} \& \bibinfo{author}{Vapnik, V.}
\newblock \bibinfo{journal}{\bibinfo{title}{Support-vector networks}}.
\newblock {\emph{\JournalTitle{Machine learning}}}
  \textbf{\bibinfo{volume}{20}}, \bibinfo{pages}{273--297}
  (\bibinfo{year}{1995}).

\bibitem{meshlab}
\bibinfo{author}{Cignoni, P.} \emph{et~al.}
\newblock \bibinfo{title}{Meshlab: an open-source mesh processing tool}.
\newblock In \emph{\bibinfo{booktitle}{Eurographics Italian chapter
  conference}}, vol. \bibinfo{volume}{2008}, \bibinfo{pages}{129--136}
  (\bibinfo{year}{2008}).

\bibitem{CT_artifacts}
\bibinfo{author}{Boas, F.~E.} \& \bibinfo{author}{Fleischmann, D.}
\newblock \bibinfo{journal}{\bibinfo{title}{Ct artifacts: causes and reduction
  techniques}}.
\newblock {\emph{\JournalTitle{Imaging in Medicine}}}
  \textbf{\bibinfo{volume}{4}}, \bibinfo{pages}{229--240}
  (\bibinfo{year}{2012}).

\end{thebibliography}

\section*{Acknowledgements}

We acknowledge the support of \textit{Shapemeans GmbH} which provided the software used to generate the statistical shape model from the 24 bones and software for fitting this model to 2D X-ray images.

\section*{Author contributions statement}

J.C. conceived the research, implemented automated data processing and network training, and conducted all experiments except for the pairwise embedding distances experiments. J.P. conducted the pairwise embedding distances experiments. P.B. provided and pre-processed the data. All the authors contributed to the manuscript. N.L., A.V. and P.B. supervised the research.

\section*{Additional information} 
Correspondence and requests for materials should be addressed to J.C.

\section*{Data and software availability}
We declare that the programming code necessary to reproduce the findings of this paper (code for data pre-processing, training the Triplet neural network and kNN classifier, and for performing inference, as well as the fully trained Triplet network and kNN classifier itself), as well as the image data (DICOM, PNG and STL files) can be requested from the corresponding author. Other deep learning models reported in this paper are publicly available in TensorFlow and on GitHub.

\end{document}